\documentclass[11pt,a4paper]{article}
\usepackage[hyperref]{emnlp-ijcnlp-2019}
\usepackage{times}
\usepackage{latexsym}
\usepackage{graphicx}
\usepackage{hhline}
\usepackage{multirow}
\usepackage{amsmath}
\usepackage{float}
\graphicspath{ {./images/} }

\usepackage{url}

\aclfinalcopy 

\title{Polly Want a Cracker: Analyzing Performance of Parroting on Paraphrase Generation Datasets}

\author{Hongren Mao, Hung-yi Lee \\
  Department of Electrical Engineering \\
  National Taiwan University \\
  {\tt ccm29cam@gmail.com, hungyilee@ntu.edu.tw} \\}

\date{}

\begin{document}
\maketitle
\begin{abstract}
  Paraphrase generation is an interesting and challenging NLP task which has numerous practical applications. In this paper, we analyze datasets commonly used for paraphrase generation research, and show that simply parroting input sentences surpasses state-of-the-art models in the literature when evaluated on standard metrics. Our findings illustrate that a model could be seemingly adept at generating paraphrases, despite only making trivial changes to the input sentence or even none at all.
\end{abstract}

\section{Introduction}

The task of paraphrase generation has many important applications in NLP. It can be used to generate adversarial examples of input text, which can then be used to train neural networks so that they become less susceptible to adversarial attack \cite{N18-1170}. For knowledge-based QA systems, a paraphrasing step can produce multiple variations of a user query and match them with knowledge base assertions, enhancing recall \cite{Yin:2015:AQC:2806416.2806542, Fader:2014:OQA:2623330.2623677}. Relation extraction can also benefit from incorporating paraphrase generation into its processing pipeline \cite{E06-1052}. Manually annotating translation references is expensive, and automatically generating references through paraphrasing has been shown to be effective for evaluation of machine translation \cite{Zhou2006ReevaluatingMT, Kauchak:2006:PAE:1220835.1220893}.

Datasets used for paraphrase generation include \textsc{Quora}\footnote{\url{https://data.quora.com/First-Quora-Dataset-Release-Question-Pairs}}, \textsc{Twitter} \cite{Lan2017ACG} and \textsc{MSCOCO} \cite{Lin2014MicrosoftCC}.
 Previous work on paraphrase generation that used these datasets \cite{SuitAndTie,Gupta2018ADG,Li2018Paraphrase,C16-1275} chose BLEU \cite{Papineni:2002:BMA:1073083.1073135}, METEOR \cite{Lavie:2007:MAM:1626355.1626389} and TER \cite{Snover06astudy} as evaluation metrics.
 
In this paper, we find that simply using the input sentence as output in an unsupervised manner ({\em i.e.} fully parroting the input) significantly outperforms the state-of-the-art on two metrics for \textsc{Twitter}, and on one metric for \textsc{Quora}. Even after changing part of the input sentence ({\em i.e.} partially parroting the input), state-of-the-art metric scores can still be surpassed.

Consequently, for future paraphrase generation research which achieve good evaluation scores, we suggest investigating whether their methods or models act differently from simple parroting behavior.

\section{Method Description}

Given an input sentence \(i\), the goal of paraphrase generation is to generate an output sentence \(o\) which is semantically identical to \(i\), but contain variations in lexicon or syntax.
Full parroting simply uses the input as output (\(o = i\)). 

Paraphrase generation models may not parrot the input sentence word for word, but it is possible that they only modify a few words of the input, thus we also experiment with simple methods of modifying \(i\), such as replacing or cutting words from the head, from the tail or from random positions.

Both full parroting and the forms of partial parroting we use are fully unsupervised.

\section{Datasets}

\textbf{\textsc{Quora.}} The \textsc{Quora} dataset contains 149,263 paraphrase sentence pairs (positive examples) and 255,027 non-paraphrase sentence pairs (negative examples). Having both positive and negative examples makes it appealing for research on paraphrase generation \cite{Gupta2018ADG, Li2018Paraphrase} and identification \cite{lan2018toolkit}.
After processing the dataset, there are 149,650 unique sentences that have reference paraphrases.

Gupta et al.~\shortcite{Gupta2018ADG} sampled 4K sentences as their test set, but did not specify which sentences they used. Li et al.\shortcite{Li2018Paraphrase} sampled 30K sentences as their test set, also not specifying which sentences they used. To avoid selecting a subset of data that is biased in favor of our method, we perform evaluation on the entire \textsc{Quora} dataset. Although we evaluate on the entire dataset, the size of our training set is zero due to the fully unsupervised nature of full and partial parroting. 

We group sentences by the number of reference paraphrases they have, and plot the relative counts in Appendix \ref{appendix:num_refs}.
It can be seen that over 64\% of entries have only a single reference paraphrase, which is problematic because even if a paraphrase of good quality is generated for any one of these entries, BLEU, METEOR and TER scores could still be inferior if the generated paraphrase differs too much from the single reference paraphrase. Previous paraphrase generation work on \textsc{Quora} \cite{Gupta2018ADG, Li2018Paraphrase} did not mention removing these entries, thus we include them in our experiments for fair comparison. However, we strongly recommend future work which wishes to use BLEU, METEOR and TER as evaluation metrics to only consider entries that have multiple reference paraphrases.

\textbf{\textsc{Twitter.}} There are 114,025 paraphrase sentence pairs in \textsc{Twitter}, which were acquired by collecting tweets which contain identical URLs \cite{Lan2017ACG}. As with \textsc{Quora}, prior paraphrase generation work on this dataset \cite{Li2018Paraphrase} did not provide their sampled test set sentences, so we evaluate parroting on the entire dataset to avoid bias. We follow the same data processing steps as \textsc{Quora}, and plot the number of reference paraphrases in Appendix \ref{appendix:num_refs}.


\textbf{\textsc{MSCOCO.}} This is an image captioning dataset, with multiple captions provided for a single image \cite{Lin2014MicrosoftCC}. There have been multiple works which use it as a paraphrase generation dataset by treating captions of the same image as paraphrases \cite{SuitAndTie,Gupta2018ADG,C16-1275}. The training and testing sets are available, containing 331,163 and 162,016 input sentences respectively.

However, relevance scores for captions of the same image score only 3.38 out of 5 under human evaluation (in contrast, the score is 4.82 for \textsc{Quora})  \cite{Gupta2018ADG}, due to the fact that different captions for the same image often vary in the semantic information conveyed. This makes the use of \textsc{MSCOCO} as a paraphrase generation dataset questionable.

We plot the number of reference paraphrases in Appendix \ref{appendix:num_refs}.

\section{Experiments}

\begin{table*}[!htbp]
\begin{center}
  \begin{tabular}{|c||c|c|c|c|c|c|}
    \hline
    \multicolumn{1}{ |c|| }{} &
    \multicolumn{3}{ |c|| }{\textsc{\textbf{STATE-OF-THE-ART}}}
    & \multicolumn{3}{ |c| }{\textsc{\textbf{PARROT}}} \\
    \cline{2-7}
    \textbf{Metric} & paper & score & \multicolumn{1}{|c||}{num\_train} & score & num\_train & $\Delta$SOTA \\
    \hhline{|=|=|=|=|=|=|=|}
    BLEU $\uparrow$ & \cite{Li2018Paraphrase} & \textbf{43.54} & \multicolumn{1}{|c||}{100K} & 41.59 & 0 & -4.47\% \\
    \hline
    METEOR $\uparrow$ & \cite{Gupta2018ADG} & 33.6 & \multicolumn{1}{|c||}{150K} & \textbf{38.60} & 0 & \textbf{+14.88\%} \\
    \hline
    TER $\downarrow$ & \cite{Gupta2018ADG} & \textbf{39.5} & \multicolumn{1}{|c||}{150K} & 45.22 & 0 & +14.47\% \\
    \hline
  \end{tabular}
  \caption{Performance of full parroting v.s. state-of-the-art on \textsc{Quora}. Higher BLEU and METEOR scores are better, while higher TER scores are worse. Bold text represents best results.}
  \label{table:quora_performance}
\end{center}
\end{table*}

\begin{table*}[!htbp]
\begin{center}
  \begin{tabular}{|c||c|c|c|c|c|c|}
    \hline
    \multicolumn{1}{ |c|| }{} &
    \multicolumn{3}{ |c|| }{\textsc{\textbf{STATE-OF-THE-ART}}}
    & \multicolumn{3}{ |c| }{\textsc{\textbf{PARROT}}} \\
    \cline{2-7}
    \textbf{Metric} & paper & score & \multicolumn{1}{|c||}{num\_train} & score & num\_train & $\Delta$SOTA \\
    \hhline{|=|=|=|=|=|=|=|}
    BLEU $\uparrow$ & \cite{Li2018Paraphrase} & 45.74 & \multicolumn{1}{|c||}{110K} & \textbf{65.26} & 0 & \textbf{+42.67\%} \\
    \hline
    METEOR $\uparrow$ & \cite{Li2018Paraphrase} & 20.18 & \multicolumn{1}{|c||}{110K} & \textbf{41.73} & 0 & \textbf{+106.77\%} \\
    \hline
    TER $\downarrow$ & - & - & \multicolumn{1}{|c||}{-} & \textbf{41.87} & 0 & - \\
    \hline
  \end{tabular}
  \caption{Performance of full parroting v.s. state-of-the-art on \textsc{Twitter}.}
  \label{table:twitter_performance}
\end{center}
\end{table*}

\begin{table*}[!htbp]
\begin{center}
  \begin{tabular}{|c||c|c|c|c|c|c|}
    \hline
    \multicolumn{1}{ |c|| }{} &
    \multicolumn{3}{ |c|| }{\textsc{\textbf{STATE-OF-THE-ART}}}
    & \multicolumn{3}{ |c| }{\textsc{\textbf{PARROT}}} \\
    \cline{2-7}
    \textbf{Metric} & paper & score & \multicolumn{1}{|c||}{num\_train} & score & num\_train & $\Delta$SOTA \\
    \hhline{|=|=|=|=|=|=|=|}
    BLEU $\uparrow$ & \cite{SuitAndTie} & \textbf{44.0} & \multicolumn{1}{|c||}{331K} & 19.0 & 0 & -56.8\% \\
    \hline
    METEOR $\uparrow$ & \cite{SuitAndTie} & \textbf{34.7} & \multicolumn{1}{|c||}{331K} & 23.9 & 0 & -31.0\% \\
    \hline
    TER $\downarrow$ & \cite{SuitAndTie} & \textbf{37.1} & \multicolumn{1}{|c||}{331K} & 70.4 & 0 & +89.7\% \\
    \hline
  \end{tabular}
  \caption{Performance of full parroting v.s. state-of-the-art on \textsc{MSCOCO}.}
  \label{table:mscoco_performance}
\end{center}
\end{table*}

We evaluate the performance of full parroting on all three datasets and compare with state-of-the-art models.

We also study the performance of partial parroting. Whereas full parroting does not modify the input sentence, partial parroting replaces or cuts some of the input words. We try three different modes of choosing words to be cut or replaced: from the sentence head, from the tail or sampled randomly.

\subsection{Evaluation}
Following prior paraphrase generation research which used \textsc{Quora}, \textsc{Twitter} and \textsc{MSCOCO}, we use BLEU, METEOR and TER as evaluation metrics. When calculating metric scores, all available reference paraphrases for a given input sentence are considered.

\subsection{Results}

\textbf{Full parroting.} Our results are organized in Tables \ref{table:quora_performance}, \ref{table:twitter_performance} and \ref{table:mscoco_performance}. We see for \textsc{Twitter}, parroting outperforms the state-of-the-art by significant margins on both BLEU and METEOR scores; for \textsc{Quora}, parroting outperforms the state-of-the-art appreciably on METEOR while having comparable performance on BLEU.

The poor performance of full parroting on \textsc{MSCOCO} is due to higher edit distances between input sentences and their reference paraphrases. TER measures the edit distance of a sentence to a reference sentence, normalized by the average length of all references \cite{Snover06astudy}:
\[\text{TER} = \frac{\text{\# of edits}}{\text{average \# of reference words}} \]

We see that the TER score of full parroting is particularly high on \textsc{MSCOCO} compared to the other two datasets. Correspondingly, the BLEU and METEOR scores are lower by a wide margin.

\begin{figure}[t]
	\centering
	\includegraphics[width=0.45\textwidth]{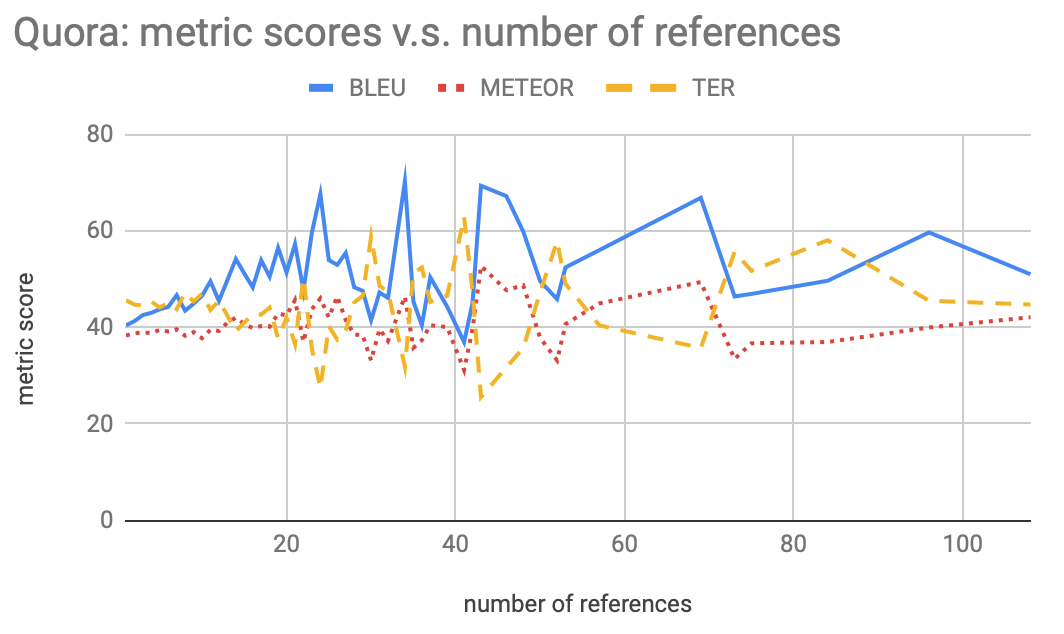}
	\caption{Metric scores v.s. number of reference paraphrases (Quora). For lower numbers of references, metric scores improve as the amount of references increases. For higher numbers of references, there does not appear to be a clear correlation.}
	\label{fig:quora_metrics_vs_num_refs}
\end{figure}
\begin{figure}[t]
	\centering
	\includegraphics[width=0.45\textwidth]{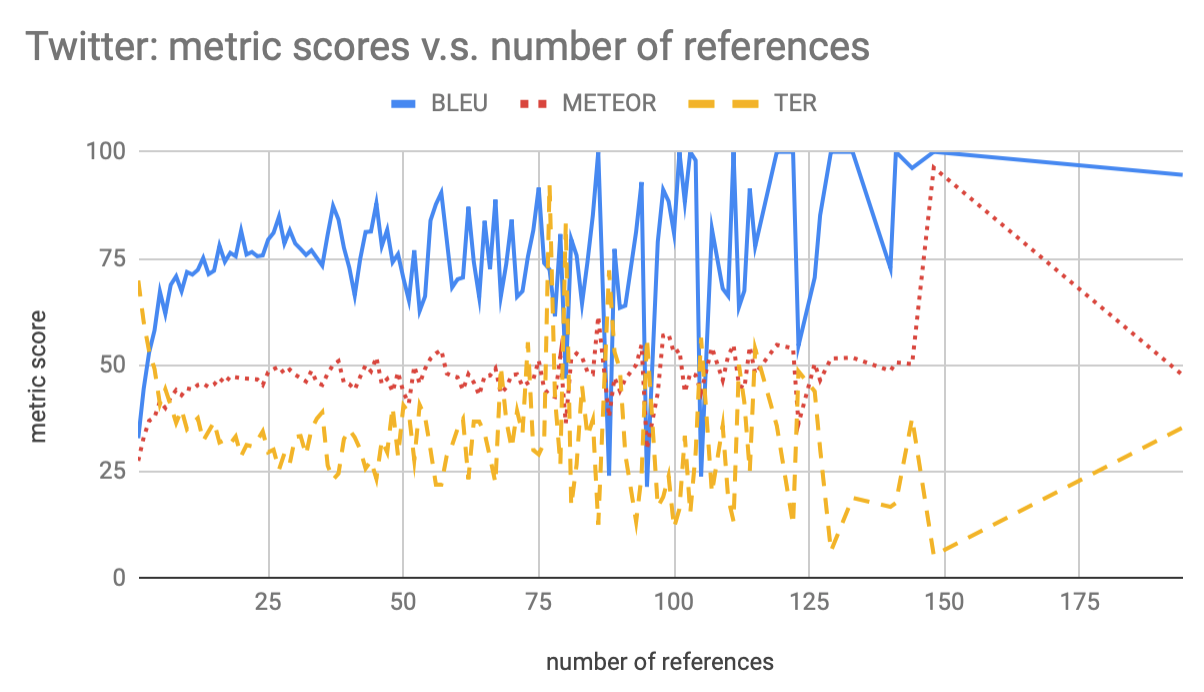}
	\caption{Metric scores v.s. number of reference paraphrases (Twitter). For lower numbers of references, metric scores improve as the amount of references increases. For higher numbers of references, there does not appear to be a clear correlation.}
	\label{fig:twitter_metrics_vs_num_refs}
\end{figure}

For further investigation of parroting on \textsc{Quora} and \textsc{Twitter}, we plot parroting performance versus the number of reference paraphrases available for a given input sentence (Figures \ref{fig:quora_metrics_vs_num_refs} and \ref{fig:twitter_metrics_vs_num_refs}). If the number of references is not too high, metric scores generally improve when the number of references rises. Once the number of references exceeds a certain threshold, we do not observe a clear correlation, showing that the probability of finding a reference sentence which bears higher resemblance to the input does not increase proportionally with the number of references.

\begin{table*}[!htbp]
\begin{center}
  \begin{tabular}{|c||c|c|c|c|c|c|}
    \hline
    \multicolumn{1}{ |c|| }{} &
    \multicolumn{3}{ |c|| }{\textsc{\textbf{Quora}} (5K test set $\times$ 1200)}
    & \multicolumn{3}{ |c| }{\textsc{\textbf{Twitter}} (4K test set $\times$ 250)} \\
    \cline{2-7}
    \textbf{Statistic} & BLEU $\uparrow$ & METEOR $\uparrow$ & \multicolumn{1}{|c||}{TER $\downarrow$} & BLEU $\uparrow$ & METEOR $\uparrow$ & TER $\downarrow$ \\
    \hhline{|=|=|=|=|=|=|=|}
    Average & 41.57 & 38.59 & \multicolumn{1}{|c||}{45.21} & 74.97 & 46.22 & 33.43 \\
    \hline
    Std. Dev. & 0.50 & 0.20 & \multicolumn{1}{|c||}{0.50} & 0.36 & 0.14 & 0.35 \\
    \hline
    Max. & 43.12 & 39.29 & \multicolumn{1}{|c||}{46.95} & 76.01 & 46.56 & 34.31 \\
    \hline
    Min. & 39.98 & 37.85 & \multicolumn{1}{|c||}{43.53} & 74.01 & 45.89 & 32.41 \\
    \hline
  \end{tabular}
  \caption{Performance of full parroting on randomly sampled test sets. The test set size and sampling method is the same as that described in prior state-of-the-art work. Here, scores for sampled \textsc{Quora} test sets are similar to those of the full dataset, and the scores for \textsc{Twitter} test sets are better than scores achieved on the full dataset.}
  \label{table:sampled_performance}
\end{center}
\end{table*}

The choice of testing on the entire dataset for \textsc{Quora} and \textsc{Twitter} experiments was to avoid bias in favor of parroting. Nevertheless, we also randomly sampled test sets of size 4K for \textsc{Quora} in the same manner as \cite{Gupta2018ADG} (which holds the most state-of-the-art records on \textsc{Quora}) and test sets of size 5K for \textsc{Twitter} in the same manner as \cite{Li2018Paraphrase} (which holds all state-of-the-art records on \textsc{Twitter}). In total, 1200 test sets of size 4K were sampled for \textsc{Quora} and 250 test sets of size 5K were sampled for \textsc{Twitter}. Parroting performance on these sampled test sets can be found in Table \ref{table:sampled_performance}. It can be observed that the average metric scores for \textsc{Quora} are similar to the scores in Table \ref{table:quora_performance}, whereas the average scores for \textsc{Twitter} are noticeably better than those in Table \ref{table:twitter_performance}. Furthermore, the score deviation between different samples is small. Consequently, although the exact test sets used by \cite{Gupta2018ADG} and \cite{Li2018Paraphrase} are not available, it is logical to assume that parroting performance would still exceed or be on par with the state-of-the-art on those test sets.

\begin{figure}[t]
	\centering
	\includegraphics[width=0.45\textwidth]{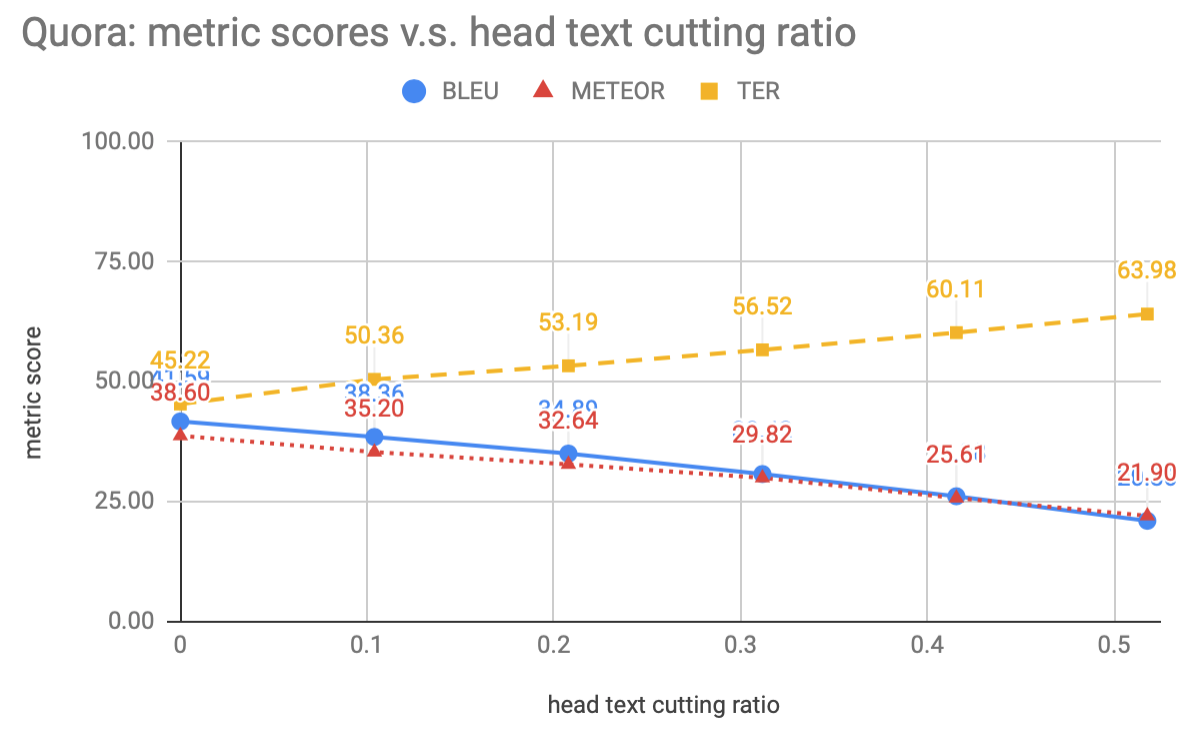}
	\caption{Metric scores v.s. ratio of text that is cut from the start of the input sentence (Quora).}
	\label{fig:quora_cut_head}
\end{figure}
\begin{figure}[t]
	\centering
	\includegraphics[width=0.45\textwidth]{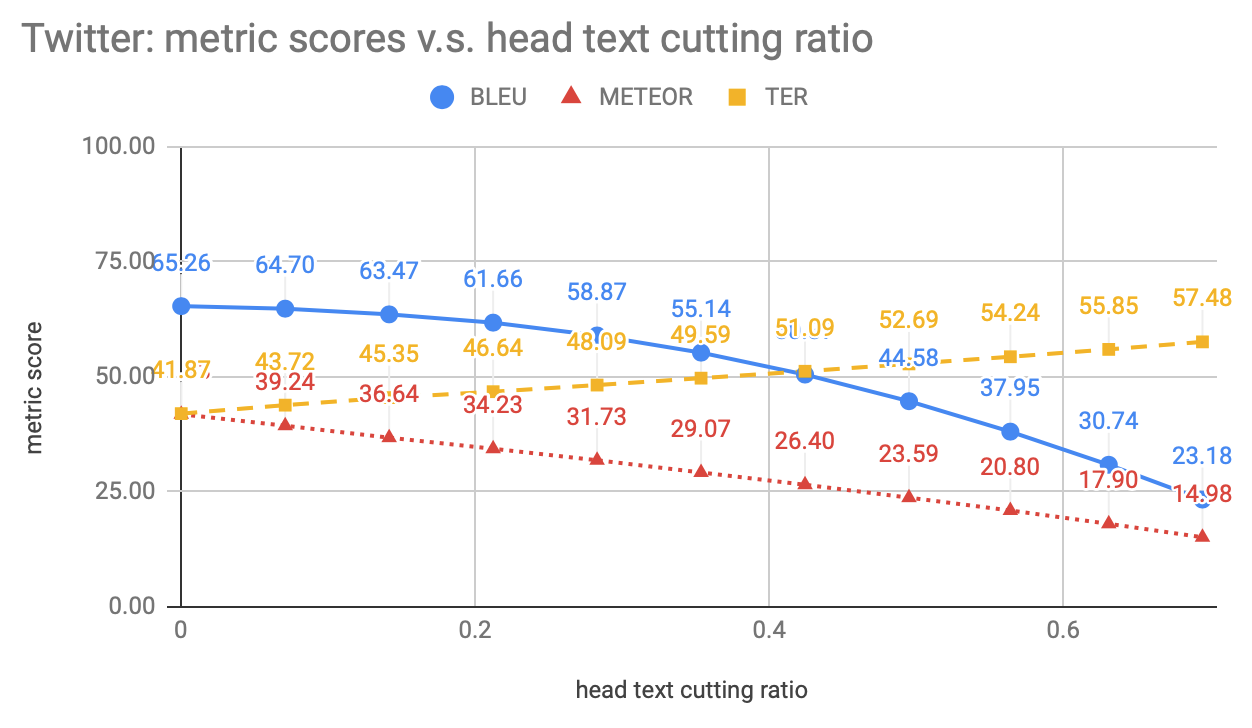}
	\caption{Metric scores v.s. ratio of text that is cut from the start of the input sentence (Twitter).}
	\label{fig:twitter_cut_head}
\end{figure}

\textbf{Partial parroting.} We also introduce lexical variation into our parroting method by replacing or cutting words of the input sentence. For replacement, we substitute input words with an out-of-vocabulary word not found in any of the input sentence's reference paraphrases. Paraphrase generation models are usually allowed to generate words which exist in reference paraphrases; we purposely use out-of-vocabulary words to give harsher scores to our method.

Figures \ref{fig:quora_cut_head} and \ref{fig:twitter_cut_head} show performance of cutting words from the start of input sentences. For \textsc{Quora}, when over 10\% of the input sentence has been modified by being cut off, partial parroting underperforms the state-of-the-art by only 3.8\% on METEOR. For \textsc{Twitter}, the same form of partial parroting (cutting off words) still outperforms the state-of-the-art on BLEU when input sentences are modified by 42\% , and does the same on METEOR when the input is modified by 56\%. Additionally, we experiment with cutting words in other positions, and also replace words rather than cut them away. The results can be found in Appendix \ref{appendix:cut_replace}.

Earlier work using \textsc{Quora} and \textsc{Twitter} \cite{Gupta2018ADG,Li2018Paraphrase} only provided a few examples of output paraphrases, and did not study in detail the paraphrasing behavior of their models, making it unclear whether the models achieve qualitatively better results than our simple rule-based parroting techniques, given that evaluation scores of the two are similar. We recommend future research to perform such an analysis if their metric scores are close to that of parroting.

\section{Related Work}

For the task of paraphrase generation, Wang et al.~\shortcite{SuitAndTie} trained a Transformer network on \textsc{MSCOCO}; Gupta et al.~\shortcite{Gupta2018ADG} trained a seq2seq variational autoencoder (VAE) on \textsc{Quora} and \textsc{MSCOCO}; Li et al.~\shortcite{Li2018Paraphrase} trained a seq2seq pointer network on \textsc{Quora} and \textsc{Twitter}, then fine-tuned it with an \textit{evaluator} which was trained via inverse reinforcement learning; Prakash et al.~\shortcite{C16-1275} trained a seq2seq model with residual connections on \textsc{MSCOCO}.

Work on paraphrase generation using other datasets can also be found. Methods include lexical substitution \cite{S07-1091, 10.1007/978-3-540-27779-8_27}, back-translation \cite{P18-1042} and sequence-to-sequence neural networks \cite{N18-1170}.

It is worth noting that paraphrase generation serves practical purposes, such as augmenting training data for NLP models to decrease their susceptibility to adversarial attack \cite{N18-1170}, or enhancing recall for QA systems \cite{Yin:2015:AQC:2806416.2806542, Fader:2014:OQA:2623330.2623677}. Improvement of downstream model performance is a valid evaluation metric for paraphrase generation, and future work wishing to use \textsc{Quora} entries which only have a single reference paraphrase could choose such an evaluation metric instead of BLEU, METEOR or TER.

As a sidenote, we also ran experiments in which BLEU scores were calculated using non-reference dataset sentences. The results are in Appendix \ref{appendix:bleu_similarity}.

\section{Conclusion}

In this work, we discover that various forms of simple parroting outperforms state-of-the-art results on \textsc{QUORA} and \textsc{TWITTER} when evaluated using BLEU and METEOR. An interpretation is that current models could simply be parroting input sentences, and researchers should perform qualitative analysis of such behavior. Another interpretation is that BLEU and METEOR are inappropriate for evaluating paraphrase generation models, in which case other metrics such as effectiveness of data augmentation \cite{N18-1170}, may be used instead.

\bibliography{emnlp-ijcnlp-2019}
\bibliographystyle{acl_natbib}

\appendix

\section{Number of References Paraphrases for Datasets}
\label{appendix:num_refs}

\begin{center}
	\centering
	\includegraphics[width=0.43\textwidth]{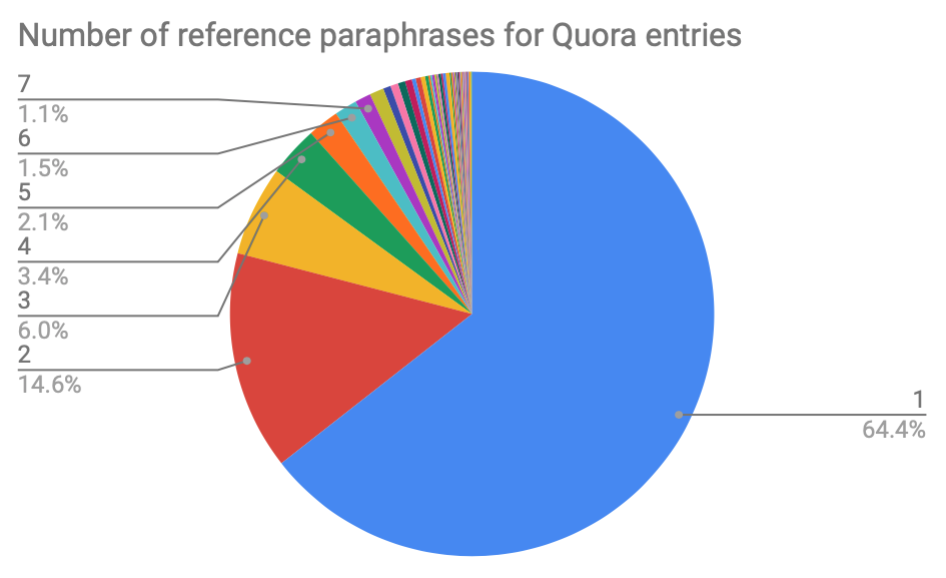}
	\captionof{figure}{Number of reference paraphrases v.s. percentage of Quora dataset}
	\label{fig:quora_num_refs}
\end{center}
\begin{center}
	\centering
	\includegraphics[width=0.45\textwidth]{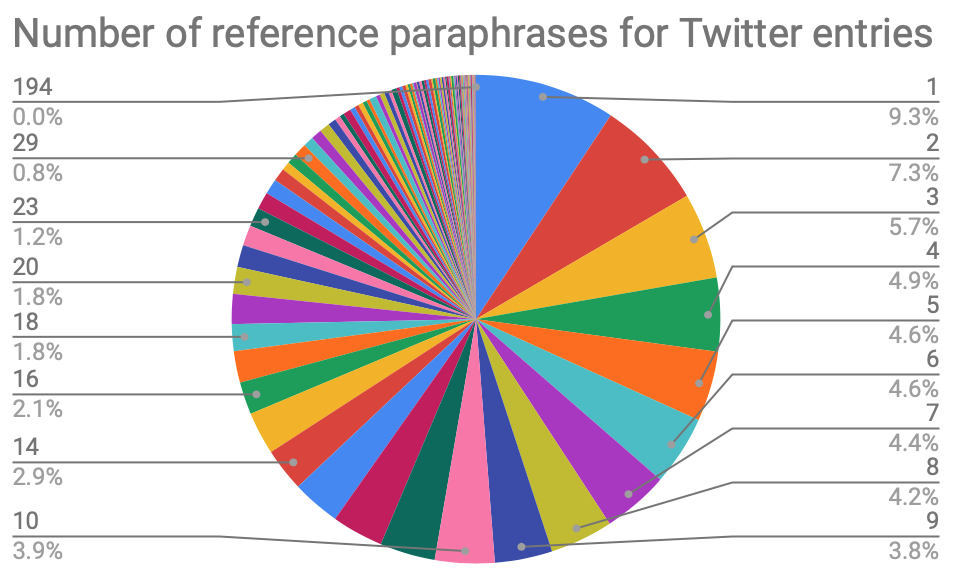}
	\captionof{figure}{Number of reference paraphrases v.s. percentage of Twitter dataset}
	\label{fig:twitter_num_refs	}
\end{center}
\begin{center}
	\centering
	\includegraphics[width=0.45\textwidth]{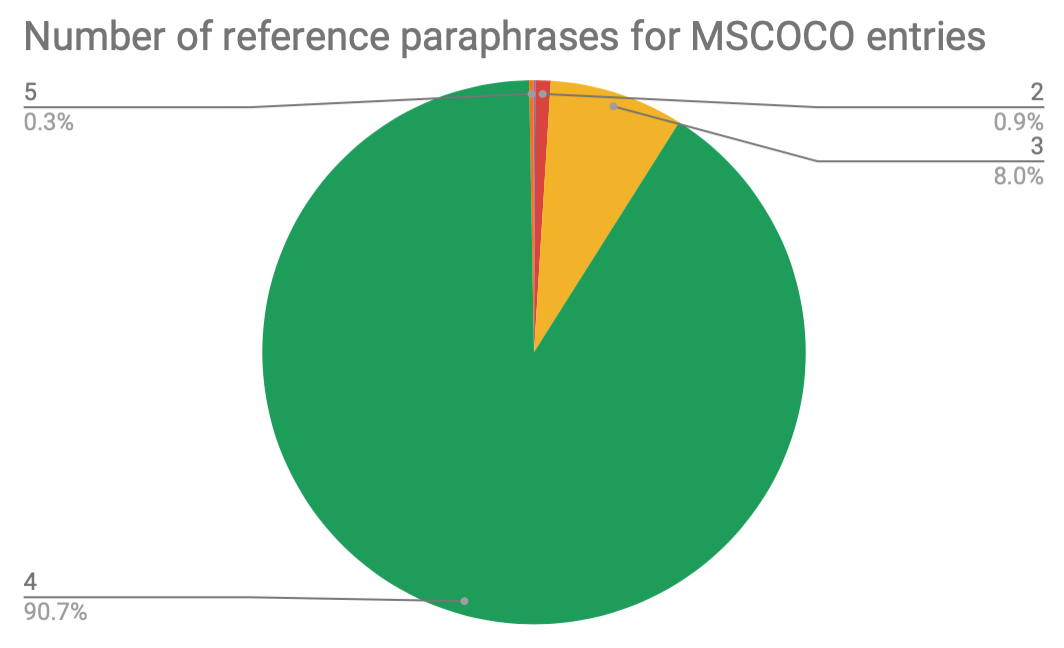}
	\captionof{figure}{Number of reference paraphrases v.s. percentage of MSCOCO dataset}
	\label{fig:mscoco_num_refs	}
\end{center}

\section{Performance of Partial Parroting}
\label{appendix:cut_replace}

\begin{center}
	\centering
	\includegraphics[width=0.45\textwidth]{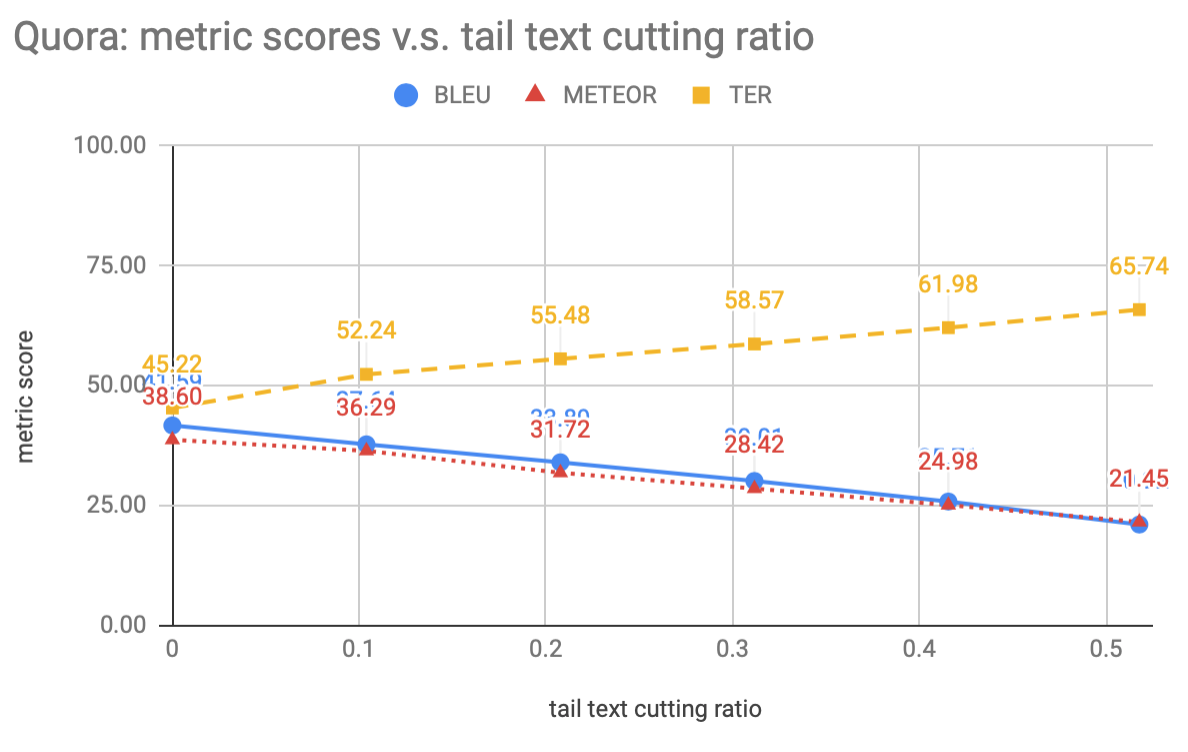}
	\captionof{figure}{Metric scores v.s. ratio of text that is cut from the end of the input sentence (Quora)}
\end{center}
\begin{center}
	\centering
	\includegraphics[width=0.45\textwidth]{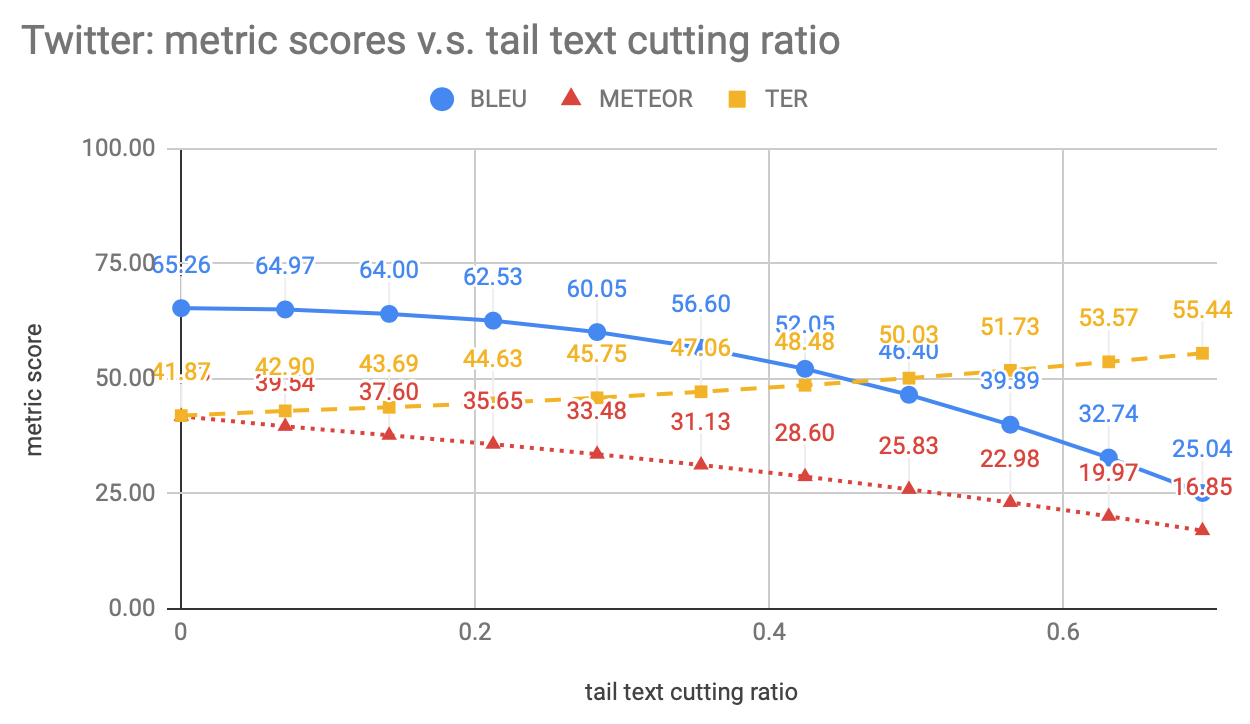}
	\captionof{figure}{Metric scores v.s. ratio of text that is cut from the end of the input sentence (Twitter)}
\end{center}
\begin{center}
	\centering
	\includegraphics[width=0.45\textwidth]{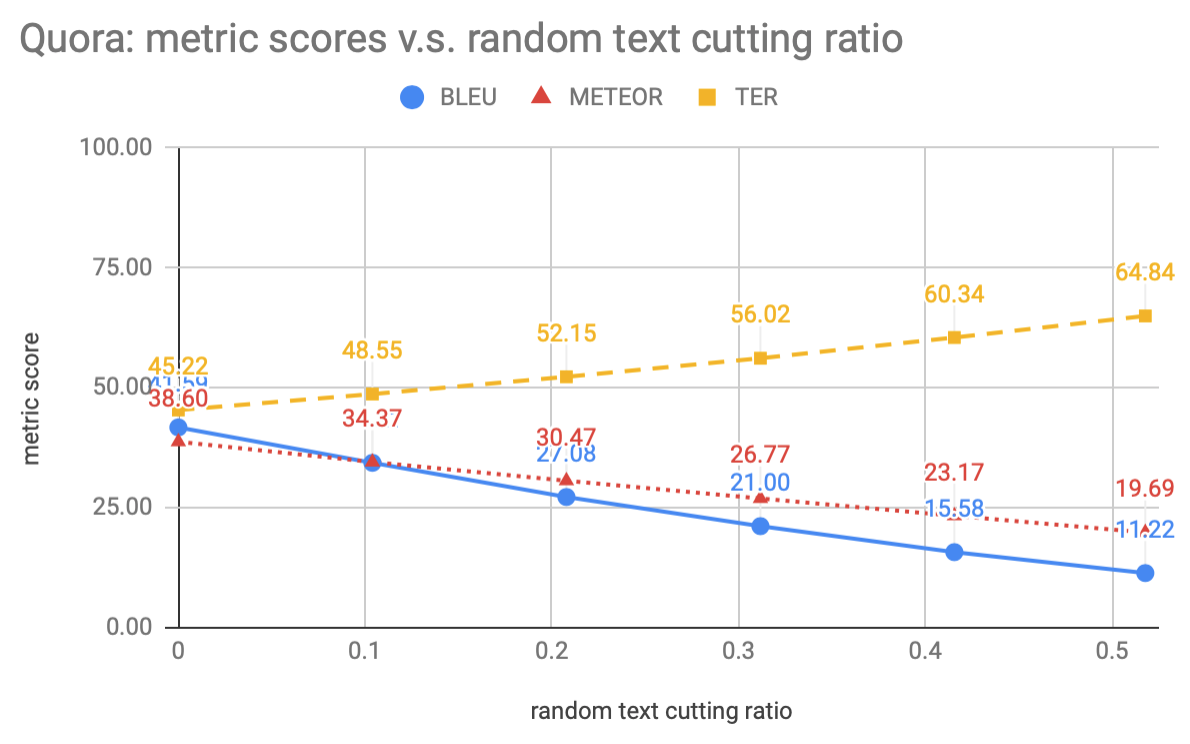}
	\captionof{figure}{Metric scores v.s. ratio of text that is cut randomly from the input sentence (Quora)}
\end{center}
\begin{center}
	\centering
	\includegraphics[width=0.45\textwidth]{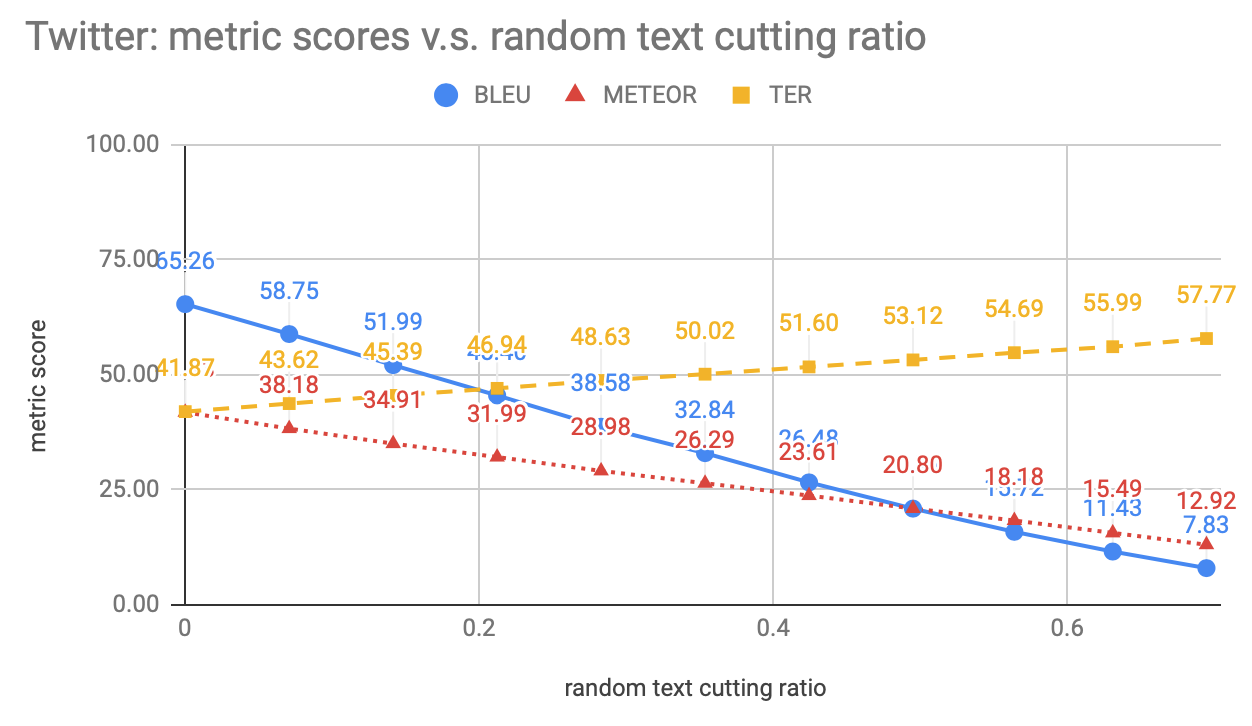}
	\captionof{figure}{Metric scores v.s. ratio of text that is cut randomly from the input sentence (Twitter)}
\end{center}
\begin{center}
	\centering
	\includegraphics[width=0.45\textwidth]{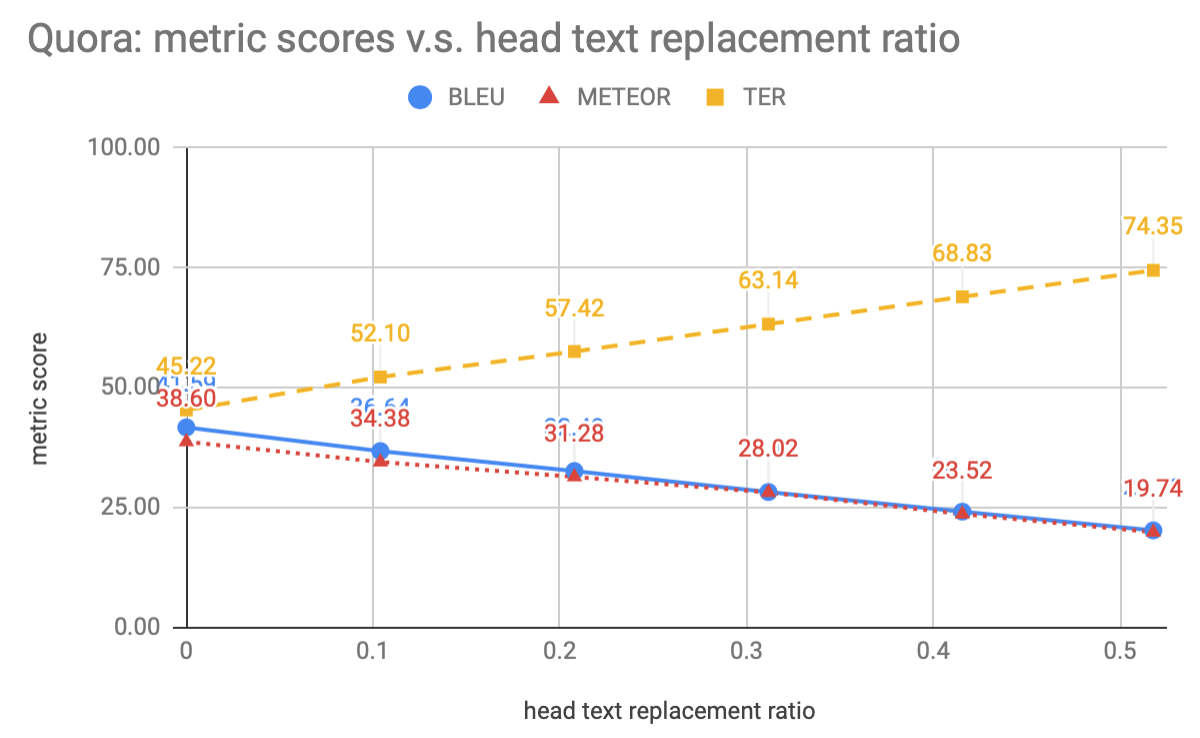}
	\captionof{figure}{Metric scores v.s. ratio of text that is replaced from the start of the input sentence (Quora)}
\end{center}

\begin{center}
	\centering
	\includegraphics[width=0.45\textwidth]{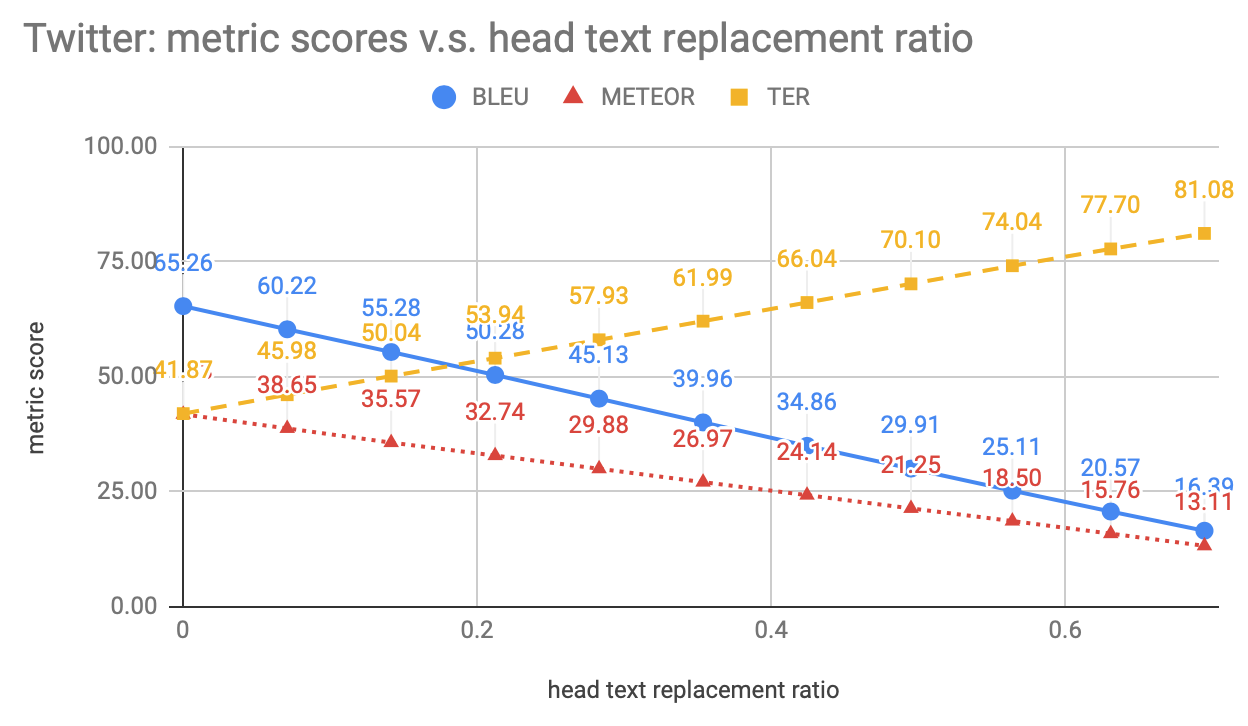}
	\captionof{figure}{Metric scores v.s. ratio of text that is replaced from the start of the input sentence (Twitter)}
\end{center}
\begin{center}
	\centering
	\includegraphics[width=0.45\textwidth]{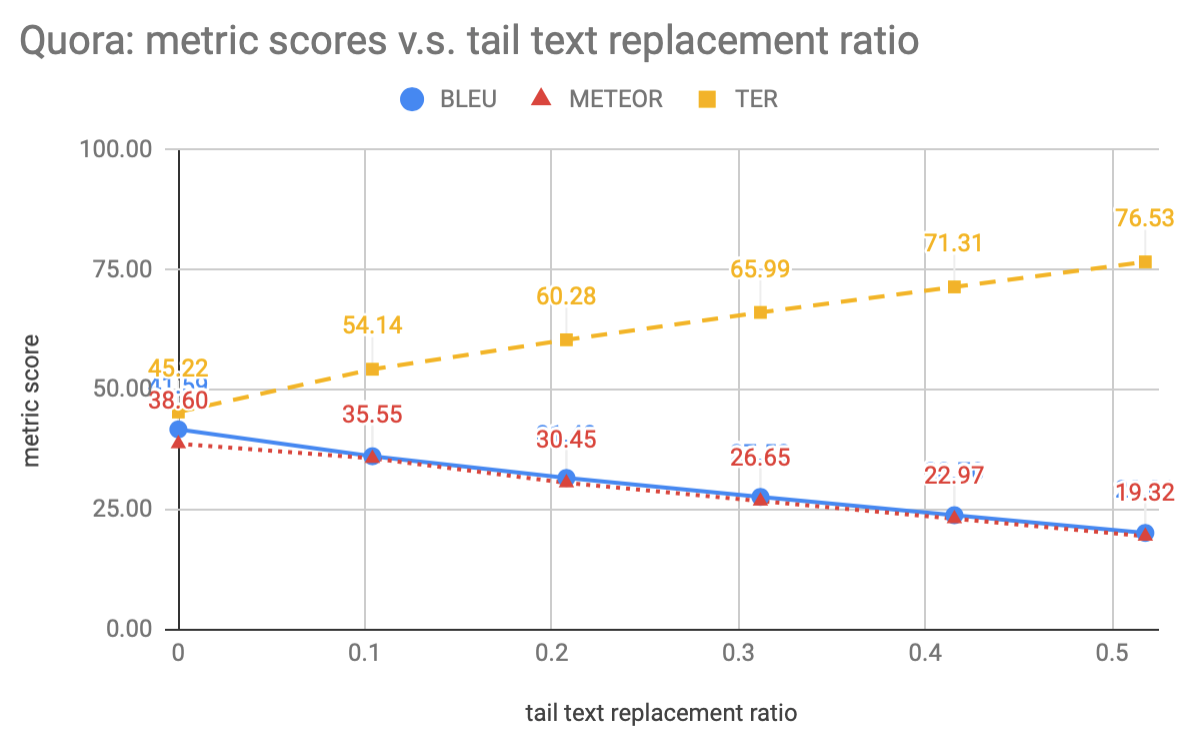}
	\captionof{figure}{Metric scores v.s. ratio of text that is replaced from the end of the input sentence (Quora)}
\end{center}
\begin{center}
	\centering
	\includegraphics[width=0.45\textwidth]{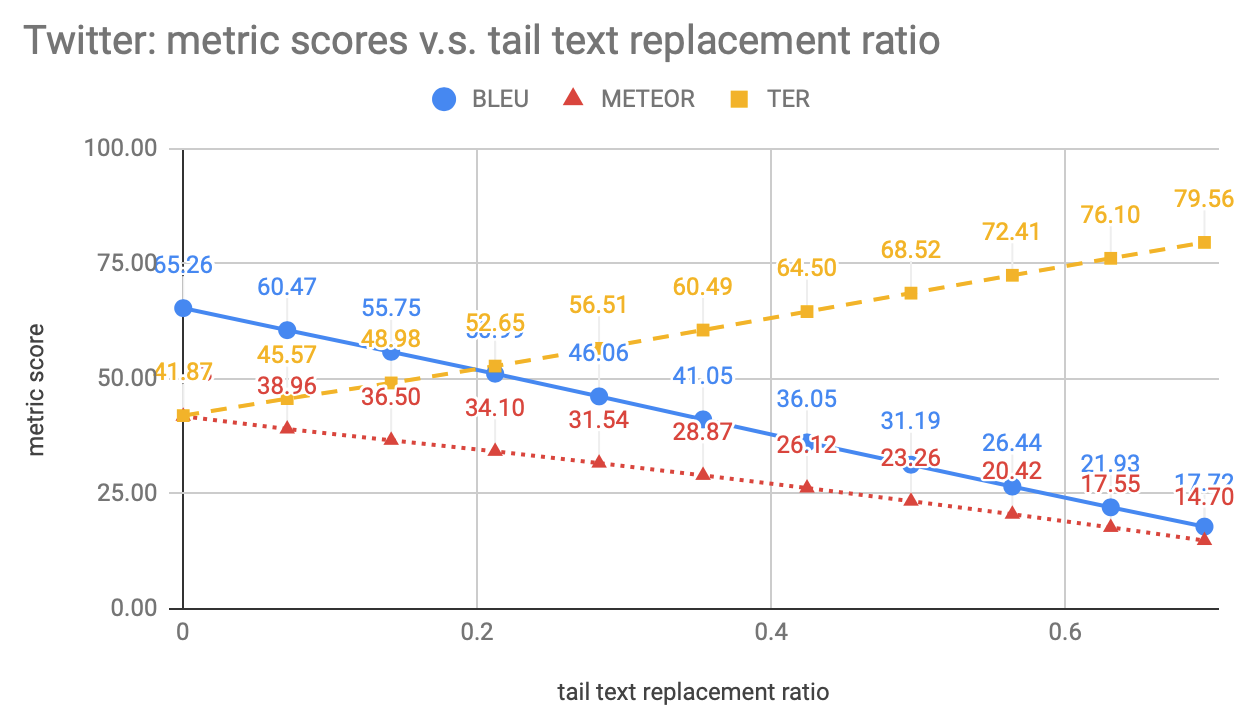}
	\captionof{figure}{Metric scores v.s. ratio of text that is replaced from the end of the input sentence (Twitter)}
\end{center}
\begin{center}
	\centering
	\includegraphics[width=0.45\textwidth]{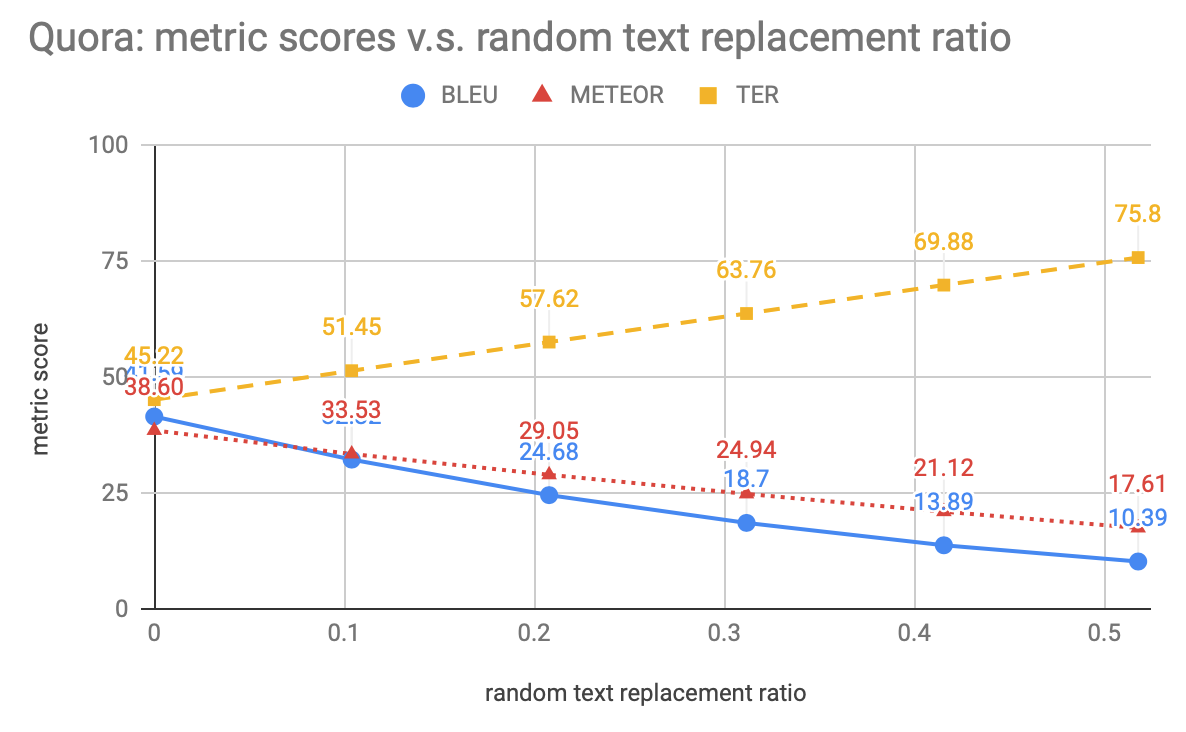}
	\captionof{figure}{Metric scores v.s. ratio of text that is replaced randomly from the input sentence (Quora)}
\end{center}
\begin{center}
	\centering
	\includegraphics[width=0.45\textwidth]{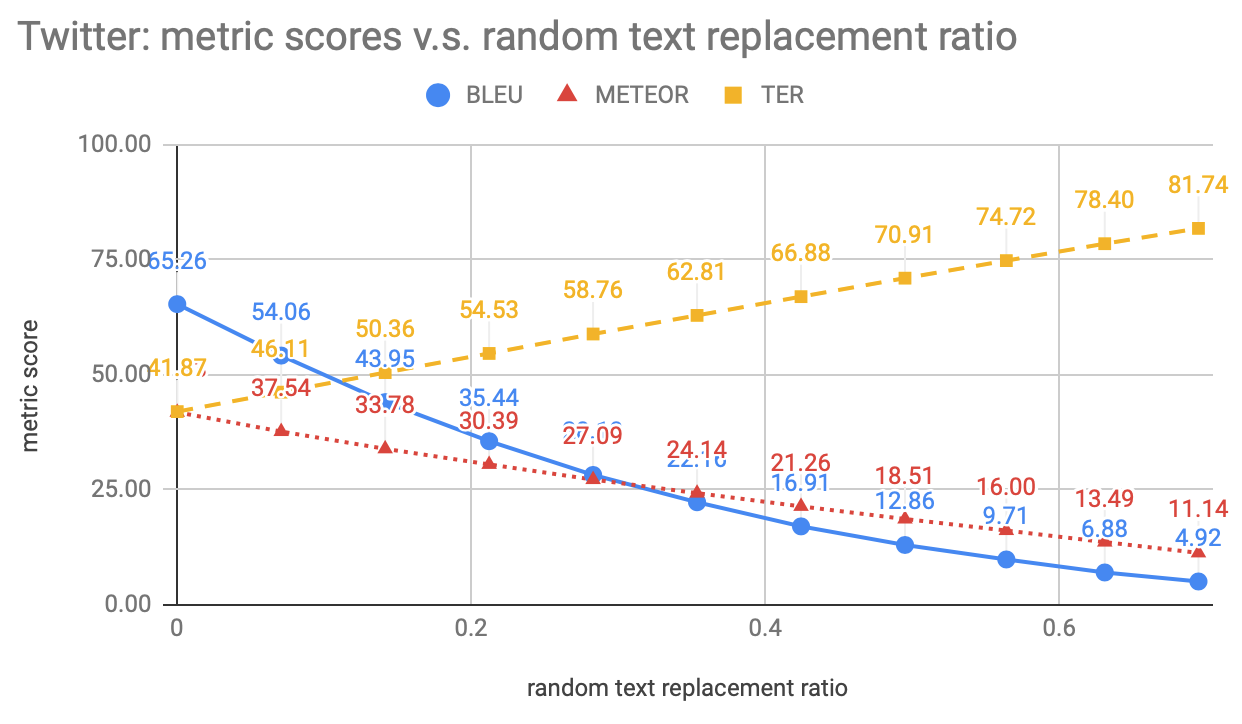}
	\captionof{figure}{Metric scores v.s. ratio of text that is replaced randomly from the input sentence (Twitter)}
\end{center}

\section{Calculating BLEU score using non-reference sentences}
\label{appendix:bleu_similarity}

For an input sentence, BLEU scores are usually calculated by comparing the input sentence with a number of reference sentences. We ran experiments in which 5 reference and 100 randomly sampled non-reference sentences were used, and show part of our results below. It can be seen that sentences with higher BLEU scores are more similar to the input sentence, which is to be expected.

\begin{table*}[p]
\begin{tabular}{lllll}
                                & Input sentence: what would happen if you hired two private detectives to spy on each other ?                                                                                                                                                                                                                                                 &  &  &  \\
\multicolumn{1}{l|}{BLEU}       &                                                                                                                                                                                                                                                                                                                                              &  &  &  \\ \cline{1-2}
\multicolumn{1}{l|}{0.3 - 0.35} & \begin{tabular}[c]{@{}l@{}}what will happen if i hire two private detectives to follow each other ?\\ what would happen if i got two private investigators to follow each other ?\\ what would happen if i sent two private investigators to find each other ?\end{tabular}                                                                   &  &  &  \\ \cline{1-2}
\multicolumn{1}{l|}{0.25 - 0.3} & (None)                                                                                                                                                                                                                                                                                                                                         &  &  &  \\ \cline{1-2}
\multicolumn{1}{l|}{0.2 - 0.25} & \begin{tabular}[c]{@{}l@{}}what would happen if earth collided with a black hole ?\\ what if i hired two private eyes and ordered them to follow each other ?\\ what would happen if donald trump lost and refused to concede the election ?\end{tabular}                                                                                    &  &  &  \\ \cline{1-2}
\multicolumn{1}{l|}{0.15 - 0.2} & \begin{tabular}[c]{@{}l@{}}would i be able to hire two private investigators and then get them to follow each other ?\\ what would happen if donald trump turned out to be a plant for hillary to win the white house ?\end{tabular}                                                                                                         &  &  &  \\ \cline{1-2}
\multicolumn{1}{l|}{0 - 0.15}   & \begin{tabular}[c]{@{}l@{}}do i need to pay again on coursera if i switch sessions ?\\ is there anyway to tell if someone blocked you on facebook ?\\ what song do you listen to when you are angry ?\\ ......\\ what are bugs you noticed on quora ?\\ if i eat a pot cookie , how long until i 'm able to pass a urine test ?\end{tabular} &  &  & 
\end{tabular}
\end{table*}

\begin{table*}[p]
\begin{tabular}{lllll}
                                & Input sentence: who do you think portrayed batman better : christian bale or ben affleck ?                                                                                                                                                                                                                                &  &  &  \\
\multicolumn{1}{l|}{BLEU}       &                                                                                                                                                                                                                                                                                                                           &  &  &  \\ \cline{1-2}
\multicolumn{1}{l|}{0.4 - 0.45} & according to you , whose batman performance was best : christian bale or ben affleck ?                                                                                                                                                                                                                                    &  &  &  \\ \cline{1-2}
\multicolumn{1}{l|}{0.35 - 0.4} & (None)                                                                                                                                                                                                                                                                                                                    &  &  &  \\ \cline{1-2}
\multicolumn{1}{l|}{0.3 - 0.35} & no fanboys please , but who was the true batman , christian bale or ben affleck ?                                                                                                                                                                                                                                         &  &  &  \\ \cline{1-2}
\multicolumn{1}{l|}{0.25 - 0.3} & (None)                                                                                                                                                                                                                                                                                                                    &  &  &  \\ \cline{1-2}
\multicolumn{1}{l|}{0.2 - 0.25} & (None)                                                                                                                                                                                                                                                                                                                    &  &  &  \\ \cline{1-2}
\multicolumn{1}{l|}{0.15 - 0.2} & \begin{tabular}[c]{@{}l@{}}what do you think about " chinese dream " ?\\ what do you think about dota2 ?\\ who was better as batman : bale or affleck ?\\ did ben affleck shine more than christian bale as batman ?\\ do you think that the demonetization policy will backfire for bjp in 2019 elections ?\end{tabular} &  &  &  \\
\multicolumn{1}{l|}{0 - 0.15}   & \begin{tabular}[c]{@{}l@{}}biswapati sarkar : how do you overcome a writer 's block ?\\ who is the better batman ? affleck or bale ?\\ how do you stop your cat from spraying ?\\ ......\\ do we always get what we deserve ?\\ can a moon have a moon ?\end{tabular}                                                     &  &  & 
\end{tabular}
\end{table*}

\end{document}